\documentclass[letterpaper, 10 pt, conference]{ieeeconf}  % Comment this line out if you need a4paper

\IEEEoverridecommandlockouts                              % This command is only needed if 
\usepackage{graphicx} % for pdf, bitmapped graphics files
\title{\LARGE \bf
RARA: Zero-shot Sim2Real Visual Navigation with Following Foreground Cues
}

\author{Klaas Kelchtermans$^{1}$ and Tinne Tuytelaars$^{1}$% <-this % stops a space
\thanks{$^{1}$KULeuven, ESAT-PSI, Belgium. {\tt\small kkelchtermans@gmail.com, tinne.tuytelaars@esat.kuleuven.be}}%
\thanks{$^{2}$Project page:
        {\tt\small github.com/kkelchte/fgbg}}%
}

\begin{document}
\maketitle
\thispagestyle{empty}
\pagestyle{empty}

%%%%%%%%%%%%%%%%%%%%%%%%%%%%%%%%%%%%%%%%%%%%%%%%%%%%%%%%%%%%%%%%%%%%%%%%%%%%%%%%
\begin{abstract} 
% 0.25p

The gap between simulation and the real-world restrains many machine learning breakthroughs in computer vision and reinforcement learning from being applicable in the real world.
In this work, we tackle this gap for the specific case of camera-based navigation, formulating it as following a visual cue in the foreground with arbitrary backgrounds.
The visual cue in the foreground can often be simulated realistically, such as a line, gate or cone.  The challenge then lies in coping with the unknown backgrounds and integrating both. As such, the goal is to train a visual agent on data captured in an empty simulated environment except for this foreground cue and test this model directly in a visually diverse real world. 
In order to bridge this big gap, we show it's crucial to combine following techniques namely:
\textit{Randomized augmentation} of the fore- and background, \textit{regularization} with both deep supervision and triplet loss and finally \textit{abstraction} of the dynamics by using waypoints rather than direct velocity commands. 
The various techniques are ablated in our experimental results both qualitatively and quantitatively finally demonstrating a successful transfer from simulation to the real world.
%Rather than solving the sim2real gap, this work intends to be an inspiration for researchers facing this huge sim2real challenge. 
Code will be made available on publication$^2$.
\end{abstract}

%%%%%%%%%%%%%%%%%%%%%%%%%%%%%%%%%%%%%%%%%%%%%%%%%%%%%%%%%%%%%%%%%%%%%%%%%%%%%%%%
\section{Introduction}
\label{intro}
% 1.5p
In 2014, Deepmind demonstrated how neural networks could reach human level performance in many Atari games by predicting controls from raw pixels \cite{atari}. 
Since then, machine learning has gained more and more impact in the robotics field.
At the same time, despite a mature control field leading to impressively natural movements as illustrated by the late Boston Dynamics spots \cite{Spot},
programming these robots still relies heavily on tedious system identification and rule-based planning.
This planning is performed in a map built by power consuming sensors, draining battery life.
In contrast, humans can perform basic navigation tasks solely relying on our eyes, building an approximate map with very low energy consumption.
Deep vision-to-control neural networks may provide us the tools to create such implicit maps based on passive camera sensors.

\begin{figure}[th]
    \centering
    \includegraphics[width=85mm]{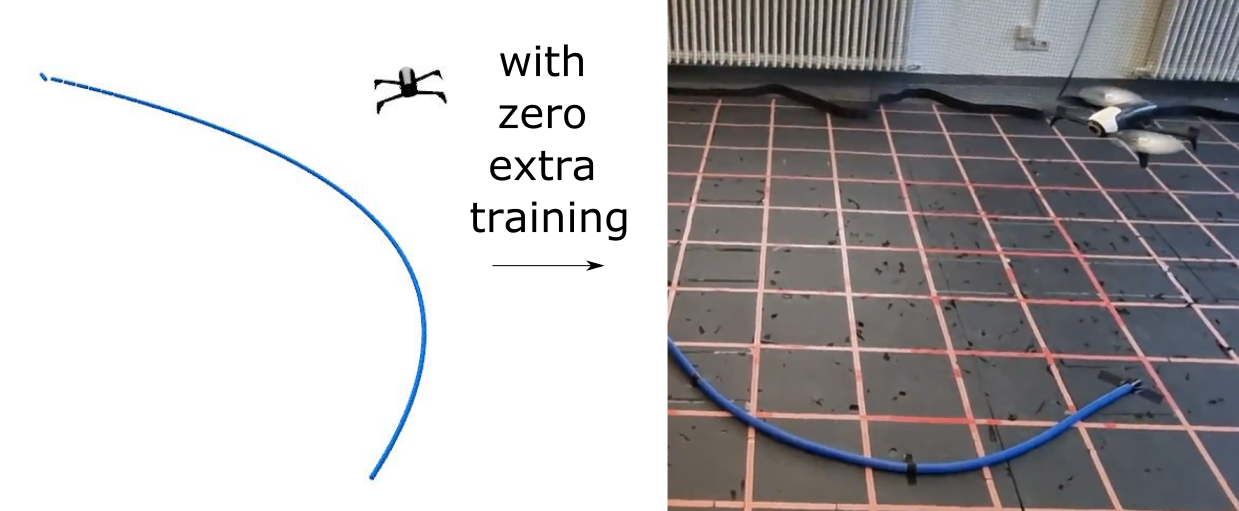}
    \caption{Demonstration of a line following behavior trained on data collected in a white simulated environment and deployed on a real Bebop drone with distracting lines in the background.}
    \label{fig:intro}
\end{figure}

Currently these vision-to-control neural networks have been demonstrated on a range of control tasks in simulation \cite{szot2021habitat}. 
The simulated environment allows parallel optimization of network parameters and automatic scenario generation for data acquisition.
Moreover, simulated environments provide a safe playground, a strict requirement for reinforcement learning methods as they are based on trial-and-error.
Unfortunately, photo-realistic simulators are not always at hand for the desired robotic task, making vision-to-control networks trained in simulation hard to deploy in the real-world.

This discrepancy between simulation and the real world has kept many deep learning methods from being applied in robotics.
On the one hand, there is a gap on the control side as the robot's dynamics will never be perfectly modelled. 
On the other hand, there is a large gap on the vision side as we have no access to a photo-realistic simulated environment.

The dynamics gap can be bypassed with the use of an abstraction~\cite{muller2018driving, kaufmann2018deep}.
Rather than predicting direct control commands, such as velocities, the neural network predicts waypoints as 3D relative coordinates.
These waypoints are sent to a controller, in our case a proportional-integral-derivative (PID), both in simulation and the real world which navigates the drone to the desired location.
In our work, we will demonstrate how this abstraction is crucial for a swift transition from the simulated environment to the real world.

The visual gap is countered in previous work with following strategies~\cite{sim2realsurvey}:
(1) Randomization: introduce visual variation in simulation so the real world is seen as just another variety~\cite{cad2rl,domainran}. 
(2) Intermediate representation: pretrain a model to extract visual information on a large real-world dataset and use the abstract representation, such as depth maps, to optimize the behavior in simulation~\cite{obstacledepth}. 
(3) Adaptation: adapt the images from the real-world to look more simplistic simulation wise, or adapt the rendered images in simulation to look more realistic, closing the visual gap between the two domains \cite{ho2021retinagan, stein2018genesis}.

In these works, either the simulator looks visually similar to the real world, making the sim2real gap not so large~\cite{cad2rl,domainran,ho2021retinagan, stein2018genesis}. 
Or the control task relies on a separate computer vision task which can be trained from a large dataset collected in the real world~\cite{obstacledepth, ho2021retinagan, stein2018genesis}.
In our work, we step aside from the assumptions of a small sim2real gap or a useful intermediate computer vision task for which real data is available. 
In contrast, we assume for the visual navigation task that there is a clear cue in the foreground indicating the desired flying direction.
This cue could be a gate, often seen with drone racing, or a lane indicated by lines or cones, or for robot grasping a specific target object. Fig. \ref{fig:intro} gives an example for line following.
Our method exploits the fact that such scene can be decomposed in foreground and background, and the sim2real gap manifests itself differently for both.
Namely, this foreground cue can be realistically modelled in simulation, however, the background is fully unknown.
As visible in Fig. \ref{fig:intro}, RGB training data is collected in an all-white environment, except for the foreground, by an expert controller with access to the desired trajectory. 
With the following techniques a neural network is trained on this data to perform the same task in a very diverse real-world setting.

These techniques in a nutshell are:
(1) Randomized augmentation over the foreground with color jitter and the background with the use of masks and a real-world places dataset.
(2) Regularization of the model's training with deep supervision and a triplet loss.
(3) Abstraction of the dynamics with the use of waypoints rather than velocities.

Summarizing, our contributions are the following:
\begin{itemize}
    \item introduction of a new approach to visual navigation with a foreground visual cue, covering a much larger visual sim2real gap than previously seen in literature,
    \item an exploring experimental study of which techniques are most suitable for this approach, combining randomization, augmentation, regularization and abstraction,
    \item a successful demonstration of sim2real zero-shot transfer to a real UAV, where each technique is ablated.
\end{itemize}

The code to reproduce our experiments is available on github$^{2}$.
Section \ref{method} gives an overview of our method.
In Section \ref{experiments} the separate parts of our method are compared on both the task of mask prediction as well as online waypoint prediction in simulation and in the real world.
Section \ref{discussion} wraps this work up with a discussion and conclusions.

%%%%%%%%%%%%%%%%%%%%%%%%%%%%%%%%%%%%%%%%%%%%%%%%%%%%%%%%%%%%%%%%%%%%%%%%%%%%%%%%
\begin{figure*}[ht]
\centering
\includegraphics[scale=0.7]{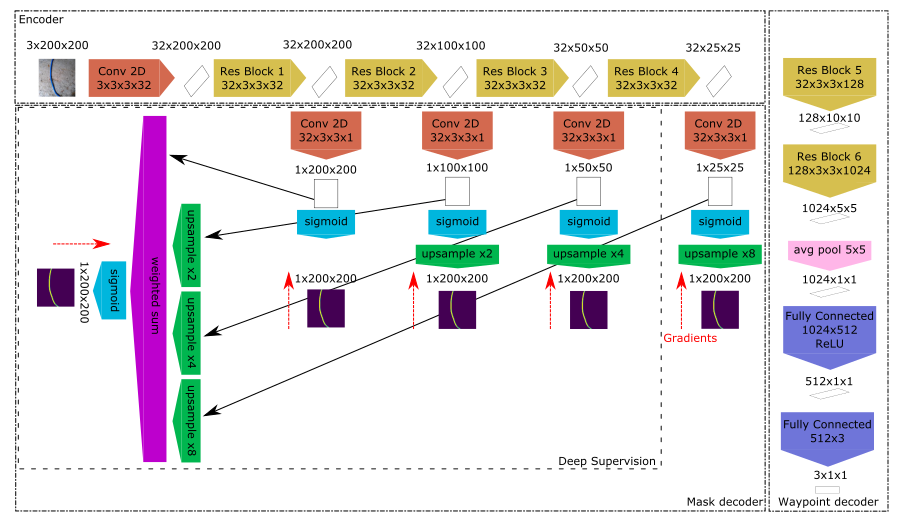}
\caption{Neural Network Architecture with the encoder at the top, the mask decoder at the bottom left and the waypoint decoder at the right. Gradients are indicated in red and masks. For the deep supervision network the combined output in the left is used as final prediction, however the loss is minimized over all mask predictions. The Res blocks follow the ResNet implementation except the identity is substituted by a convolutional layer.}
\label{fig:arch}
\end{figure*}
\section{Method}
\label{method}
% 1.5p
This section elaborates on the various techniques combined to tackle the defined visual navigation. First, it introduces the setup for data acquisition in Section \ref{setup}.
After this, the data augmentation techniques are discussed in \ref{augmentation}. Next, Section \ref{dnn} explains the neural network architecture for predicting waypoints. The last two sections introduce two regularization techniques for reducing the potential overfitting, namely deep supervision in Section \ref{deepsupervision} and the triplet loss in Section \ref{contrastivelearning}.

\subsection{Setup}
\label{setup}
% simulated environment
Visual navigation is performed on a Bebop 2 with downward facing camera interfaced with ROS on an edge computer. 
In the Gazebo simulator, a similar drone model is used to collect data by flying over a randomly generated lines.
The lines are textured with an image taken from the water tube used in the real setup.
The lines are generated by a spline matching points forming various curvatures and are build in 3D with short cylindrical tubes in Gazebo. The background is kept bright white, making the extraction of a foreground mask easy. The training set contains 5599 images.

% PID controller
Data is collected by generating a world with a line and controlling the drone with a PID controller where each line partition corresponds to a waypoint.
The PID controller is taken from \cite{mathiascontroller} and is combined with an asynchronous Kalman filter.
At each time step during data collection in simulation, the following data is stored: 
the RGB camera image $c$, the extracted mask $m$ indicating which pixels correspond to the foreground, the relative waypoint coordinates $w$ and the applied velocities $v$. 

In order to evaluate the masking performance on real-world images a small set of six images was collected on various backgrounds with various light conditions and manually labelled. As the labelling is a tedious job and the results are consistent over all six images, we decided to stick to this small number of images. This set is further referred to as the out-of-distribution set (OOD) as the images are not sampled from the same distribution as the training images.

\subsection{Randomized Augmentation of the simulated training data}
\label{augmentation}
% BG datasets
% FG variation
The goal is to use a neural network trained in simulation directly in a visually very different real-world setting.
In order to do so, the set of training data needs to be augmented in such a way that the real input is covered by the generalization capabilities of the network.
The neural network is trained on augmented input images, $I$, that combine the simulated foreground, $C$, with augmented background, $A$, using the extracted mask, $M$:
$$I = M * C + (1-M) * A$$
with $A$ a randomly selected image from a real-world dataset. Note that as only the RGB images are used and not the labels of the dataset, we are not restricted to a predefined computer vision task.

In our experiments, the background images are randomly picked from the Describable Texture Dataset \cite{dtd} with 5.7k images and MIT Places dataset \cite{places} with 36.5k images.

\begin{figure}[ht]
    \centering
    \includegraphics[width=85mm]{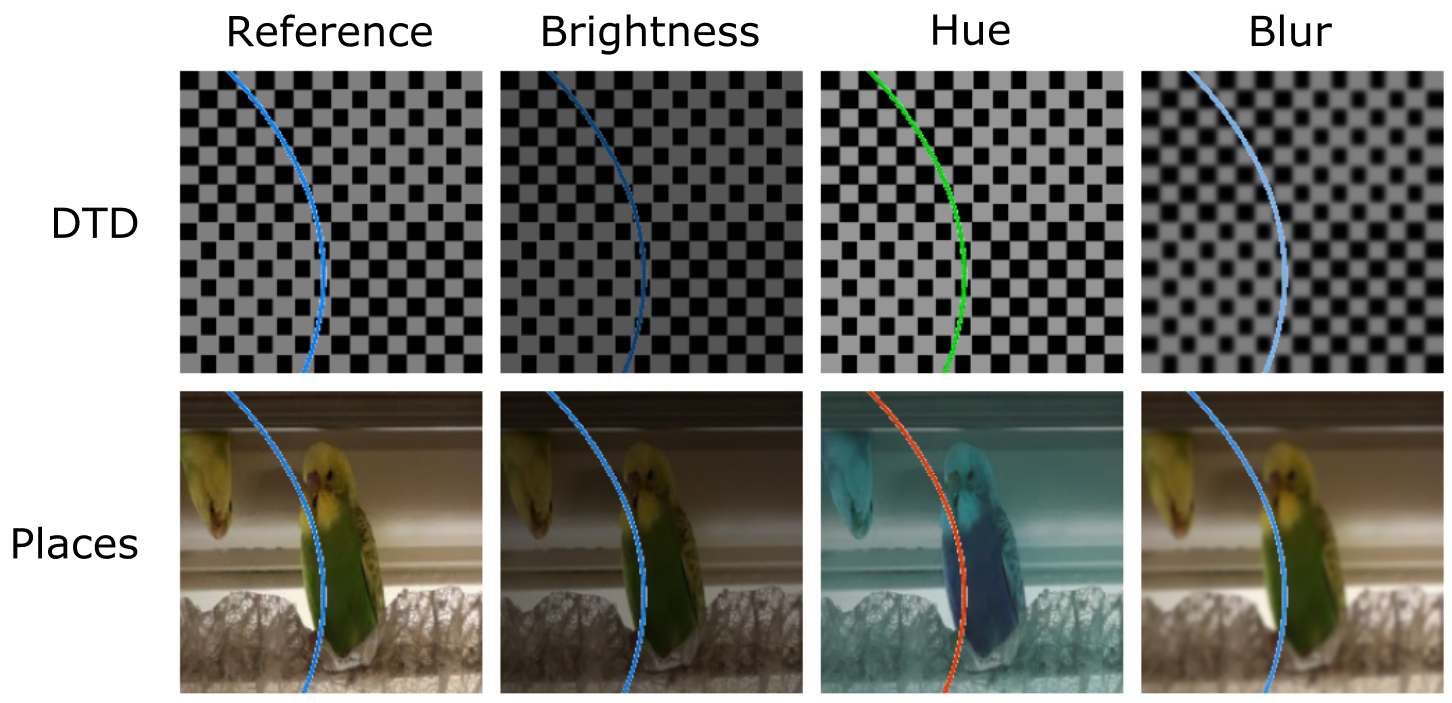}
    \caption{Example training images with backgrounds from DTD and places (row one and two) and foreground augmentation of brightness, hue and blur (column two to four).}
    \label{fig:trainingimgs}
\end{figure}

The simulated foreground can never match all possible real-world variations. Therefore a crucial extra foreground augmentation step is added in the form of color jitter \cite{colorjitter} and blur \cite{gaussianblur}. 
The brightness and hue were randomly adjusted with a factor 0.5.
In order to compensate for the motion blur, the foreground was also randomly blurred with a Gaussian blur with kernel size 9 and standard deviation sampled between 0.1 and 2.
Some resulting training images are displayed in Fig. \ref{fig:trainingimgs}.

\subsection{Abstraction: deep mask and waypoint prediction}
\label{dnn}
% RES blocks + decoder + WBCE
% WP decoder
The neural network architecture contains an encoder network with four consecutive RES blocks \cite{resnet},
each down scaling the input image from 200$\times$200 with a factor two and doubling the channel dimension, as shown in the top of Fig. \ref{fig:arch}.
The architecture has two decoders.
One decoder predicts the foreground mask, visible in the lower center block of Fig. \ref{fig:arch}, the other the 3D waypoint coordinates, visible in the right block of Fig. \ref{fig:arch}.

The mask decoder combines the feature maps into one binary mask, which is upsampled to the original resolution.
The output predicts for each input pixel whether this pixel corresponds to the visual cue in the foreground.
Both the weights of the encoder and the mask decoder are optimized with a weighted binary cross-entropy loss (WBCE):

$$
L = - \beta \sum_{(j \in M^+)} log(Pr(x_j=1))
$$
$$
- (1-\beta) \sum_{(j \in M^-)} log(Pr(x_j=0))
$$

with $M^+$ and $M^-$ the set of pixel indices corresponding to the foreground and background in the mask label, $x_j$ is the corresponding prediction and $\beta$ is a weight term to balance out the small number of foreground pixels with the large number of background pixels. For all our experiments $\beta$ was set to 0.9. For interpreting the performance, the predicted mask is rounded to a binary mask with a 0.5 threshold, allowing the calculation of an intersection over union (IoU) with the labelled mask.

The waypoint decoder, as shown on the right of Fig. \ref{fig:arch}, is trained in a second step with the encoders weights frozen. 
First the feature maps are further projected into a one dimensional vector of dimension 1024 with two RES blocks and an average pooling layer. 
This feature is decoded with two fully connected layers into a vector predicting the relative x, y, z coordinates of the next waypoint $\hat{w}$.
The decoders weights are trained with a simple mean-squared error (MSE) loss between the prediction $\hat{w}$ and the labels $w$.

Note that the final goal of the neural network is to predict reliable waypoints for the PID controller. The mask decoder can be seen as an auxiliary output used to pretrain the encoder. Thanks to the mask predictions, we can get an insight of which visual cues are detected and represented in the latent feature before it is decoded into a waypoint.

In order to ablate the abstraction of waypoints with respect to predicting velocities commands, the waypoint decoder was adjusted to predict 4D velocities $v$, of which three linear velocities in x, y, z and one angular velocity in yaw.

\subsection{Regularization I: Deep Supervision (DS)}
\label{deepsupervision}
Due to the bottleneck introduced by decoding the input image to a small feature map, the output resolution is poor.
Extracting a good mask requires global visual information with the overall layout of the line, as well as local information to get a sharp edge.
Deep supervision makes this possible by forcing each intermediate set of feature maps to represent the segmented mask as good as possible \cite{deepsupervision}.  
Each intermediate set of feature maps is combined into a mask, where all masks are combined with trainable weights to a final prediction.
Thanks to this combination, all layers have a short link with the loss layer, forcing each layer to focus on what matters most, namely the extraction of the visual cue.
Thanks to the final combination layer, both the coarse as well as the fine-grained information is available, resulting in much sharper masks.

The details of training this architecture is shown in the lower left block of Fig. \ref{fig:arch}. The red arrows indicate the gradients calculated from the WBCE, equally for each of the intermediate outputs.

\subsection{Regularization II: Triplet Loss (TL)}
\label{contrastivelearning}
From the area of contrastive learning, the triplet loss has shown to be most effective in regularizing features by pushing representations of similar inputs together and of dissimilar inputs apart \cite{tripletloss}.
The triplet loss is defined as follows, with the math borrowed from the pytorch manual \cite{tripletmarginloss}: 
$$L(a, p, n) = \max \{d(a_i, p_i) - d(a_i, n_i) + {\rm margin}, 0\}$$
with $L$ the loss based on three features: the anchor $a$, the positive or similar feature $p$ and the negative or dissimilar feature $n$. If the distance between the anchor and the similar feature is larger than the distance with the anchor and the negative feature, the loss will be non-zero. A margin term is added for stability and kept to the default value of one.

Thanks to the background augmentation, this loss can be easily applied by taking a reference RGB input from simulation, $r$, and combine it with background image, $b$. They will serve as anchor. The positive sample is created by combining $r$ with another background image $b'$. The negative sample is created by combining another foreground $r'$ with the same background image $b$. In this way, the network has an extra incentive to focus on the foreground and not being distracted by the background.

\begin{figure}[ht]
    \centering
    \includegraphics[width=80mm]{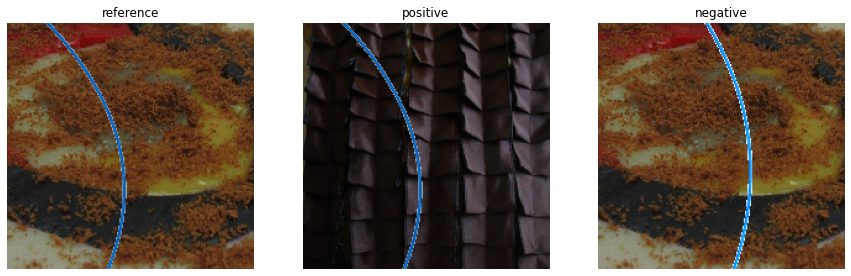}
    \caption{Example of anchor, positive and negative example for triplet loss.}
    \label{fig:tripletexample}
\end{figure}

Fig. \ref{fig:tripletexample} gives an example of augmented data triplets. The triplet loss is calculated on the one dimensional features after the average pooling layer in the waypoint decoder. This means that Res block 5 and 6 are in this case also optimized in the first training stage with the mask decoder. Also during the second stage of training the waypoint decoder, these blocks are further optimized with both the MSE and the triplet loss.

%%%%%%%%%%%%%%%%%%%%%%%%%%%%%%%%%%%%%%%%%%%%%%%%%%%%%%%%%%%%%%%%%%%%%%%%%%%%%%%%
\section{Experiments}
\label{experiments}

% 2p
Three sets of experiments are performed. First the foreground extraction methods are evaluated quantitatively and qualitatively on their capabilities to transfer directly to real-world data in Section \ref{maskpred}. Section \ref{wpsim} describes the experiments of the waypoint and velocity prediction in simulation. In Section \ref{wpreal} the real-world experiments are described as well as their results.

\subsection{Mask prediction}
\label{maskpred}

\begin{table}[ht]
\caption{Mask Prediction performance on validation and OOD data with standard deviation between brackets}
\label{tab:mask}
\begin{center}
\begin{tabular}{r|c|c}

Network & validation IoU & OOD IoU \\
\hline
Vanilla & 34.38 (1.0) & 34.95 (13) \\
\hline
Places BG & 34.18 (0.9) & \textbf{45.74} (22) \\
DTD BG & 33.99 (0.9) & 36.75 (26) \\
DTD and places BG & 34.20 (0.9) & 38.44 (20) \\
\hline
Brightness FG & 34.20 (0.8) & \textbf{53.53} (8) \\
% Contrast FG & 34.28 (0.8) & 33.89 (19) \\
Hue FG & 34.11 (1.0) & 23.63 (21) \\
% Saturation FG & 34.23 (0.9) & 45.41 (15) \\
Blur FG & 34.46 (1.1) & 33.52 (17)\\
\hline
DS & 89.83 (0.2) & 39.62 (14) \\
DS with TL & 87.66 (0.2) & \textbf{53.69} (11) \\

\end{tabular}
\end{center}
\end{table}

In this section, the various techniques described above are evaluated on the task of line following. Table \ref{tab:mask} gives an overview with both the average validation IoU and out-of-distribution IoU. The validation data is taken from the same distribution as the training images, while the out-of-distribution data corresponds to the manually collected dataset in the real world.

The vanilla baseline is trained solely on the simulated data without background augmentation. The predicted masks on real images have a surprisingly large intersection over union of 35\%. 

Initially the background is augmented for the two separate datasets as well as combined. The places dataset reaches the highest intersection-over-union although the variation on the results is large, so this result is not significant. However, as the places dataset has a larger variety and more images, the next experiments are always augmented with this dataset.

The results of augmenting the foreground image with random color and blurring are shown from row five to row seven in Table \ref{tab:mask}. Only the variation in brightness yields a significant improvement on the out-of-distribution data, demonstrating how the simulated foreground is similar enough to the real world except for the light conditions. The next experiments will only use this brightness augmentation. This model will further be referred to as the baseline.

\begin{figure*}[ht]
    \centering
    \includegraphics[width=160mm]{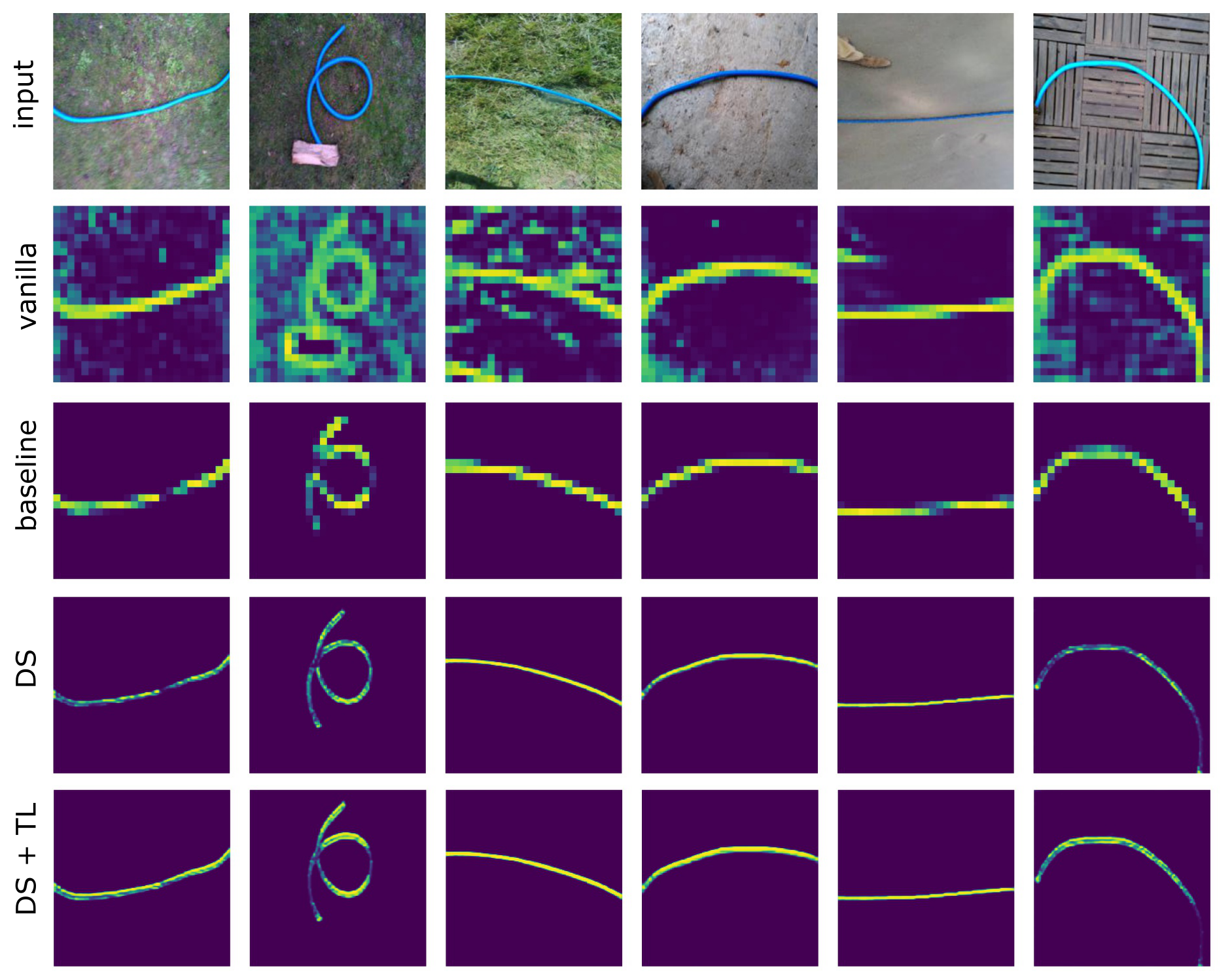}
    \caption{Qualitative results of mask prediction on out-of-distribution data}
    \label{fig:qual}
\end{figure*}

\begin{figure}[ht]
    \centering
    \includegraphics[width=85mm]{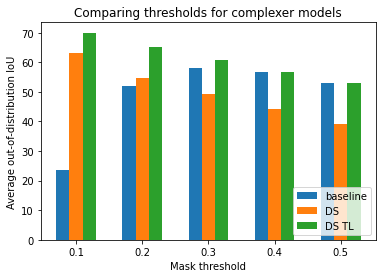}
    \caption{Performance for some smaller thresholds with DS + TL gaining up to 70\% IoU. }
    \label{fig:threshold}
\end{figure}

By introducing deep supervision as well as a triplet loss, the performance on the validation jumps from 34 IoU to 89 IoU, thanks to the higher resolution. Remarkably the out-of-distribution performance drops. This drop can be explained by the increase of complexity with deep supervision. The model is capable of fitting well to the training data and without a regularising with a triplet loss, it overfits. 
Unfortunately even with the triplet regularisation loss, the performance is not boosted significantly on the OOD data. However, when we look at the qualitative results in Fig.~\ref{fig:qual}, the quality of these predictions are much better. In order to understand what is going on, we need to take a closer look at the IoU calculation.
As the predicted mask is clipped to a binary mask with a threshold of 0.5, the regions where the network is less certain will have a smaller output value and will be clipped to zero. By varying this output threshold, the best threshold can be found. In Fig.~\ref{fig:threshold} the baseline model could reach up to 60\% IoU for a threshold of 0.3. The more complex models however get up to a 70\% IoU thanks to the improved resolution. Also the positive impact of the regularisation term (TL) is clearly visible. 
However, when the threshold is set too high, the negative impact of the 'uncertainty' of the deep supervision is clear with a performance drop. By adding the triplet loss (TL) the negative impact is compensated.

\subsection{Waypoint prediction in simulation}
\label{wpsim}

\begin{figure}[ht]
    \centering
    \includegraphics[width=60mm]{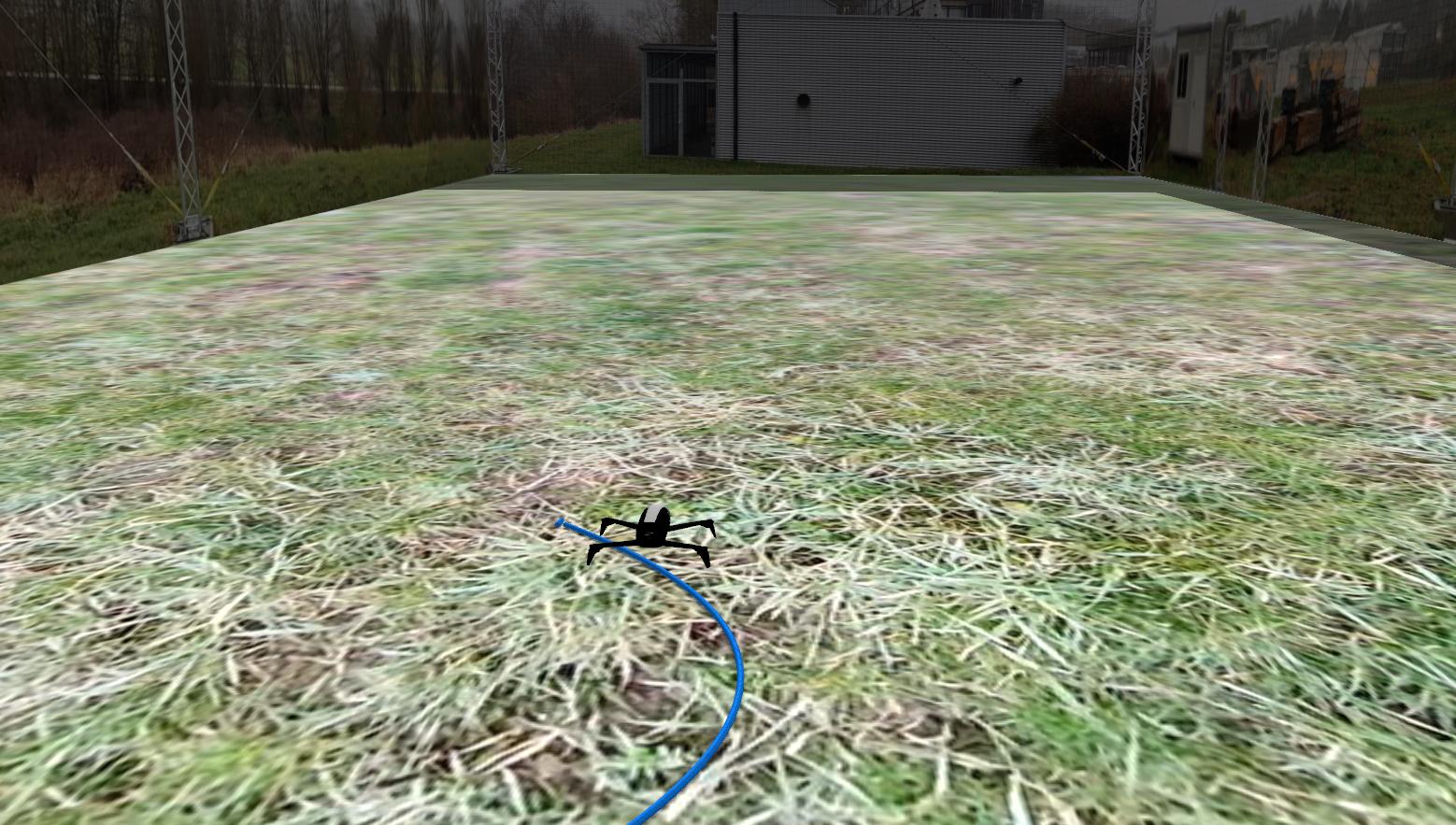}
    \caption{Out-of-distribution simulated environment for online waypoint prediction evaluation.}
    \label{fig:wpsim}
\end{figure}

In a second set of experiments the actual waypoint prediction is evaluated in a more realistic simulated environment, shown in Fig. \ref{fig:wpsim}. Ten times a line is randomly spawned and a PID controller is used to navigate the drone based on the predicted waypoints. The results are summarized in Table \ref{tab:wpsim}, showing the offline validation RMSE at the end of training, the online RMSE averaged over the ten runs and the success rate. All models were successful at navigating 10 times over the defined line of 2m length without loosing track. The online RMSE of the baseline model is already very good (1.5cm). Deep supervision does not seem to be that beneficial even with the triplet loss added. As the results are only half a standard deviation apart, the performance appear quantitatively very similar.

In order to evaluate the abstraction with waypoints versus direct prediction of command velocities, the DS + TL model was also trained with a velocity decoder predicting 4D velocity commands (x, y, z and yaw) instead of 3D waypoint coordinates. The last row of Table \ref{tab:wpsim} shows this model converges to a much larger RMSE both when switching from the offline validation data to the online interactive setting. This shift from offline to online data is referred to as the state space shift~\cite{ross2014reinforcement}. In general the shift was small enough to ensure a reasonable behavior and so a 10/10 success. 

\begin{table}[ht]
    \centering
    \begin{tabular}{l|c|c|c}
             & Validation     & Online  & Success \\
             & RMSE  $[m]$ & RMSE  $[m]$ & rate \\
    \hline
    Baseline & 0.0123 (0.011) & \textbf{0.0148} (0.008) & 10 / 10 \\
    DS       & 0.0126 (0.011) & 0.0192 (0.009) & 10 / 10 \\
    DS + TL  & \textbf{0.0117} (0.009) & \textbf{0.0149} (0.008) & 10 / 10 \\
    DS + TL + W/O Abs & 0.1077* (0.093) & 0.2946* (0.25) & 10 / 10 \\ 
    \end{tabular}
    \caption{Comparing offline and online waypoint prediction errors and success rate. *Without abstraction: RMSE of 4D velocity commands instead of 3D waypoints. These values are indicative for completeness but should not be compared directly with the waypoint errors.}
    \label{tab:wpsim}
\end{table}

\subsection{Waypoint prediction in the real world}
\label{wpreal}

\begin{figure*}[ht]
    \centering
    \includegraphics[width=170mm]{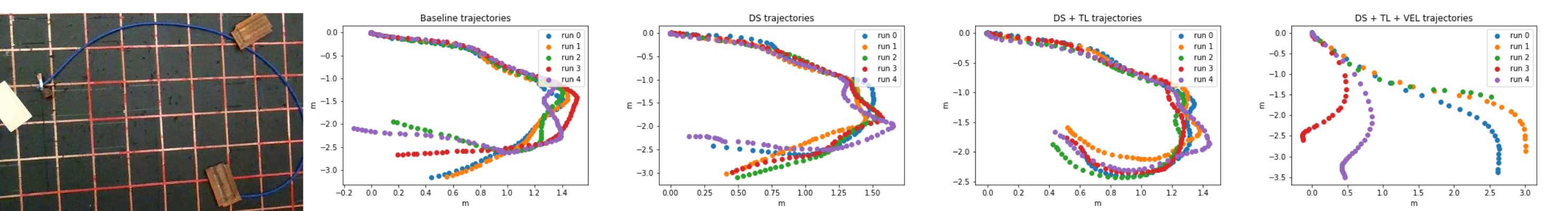}
    \caption{Real-world experiment setting with top down view of five trajectories for each model, starting at (0, 0), indicated with the white board. Only the best DS + TL model is capable of not loosing track. Using direct velocity commands instead of waypoints lead to poor results as well.}
    \label{fig:wpreal}
\end{figure*}

Finally the models are evaluated five times in a real-world scenario. 
With edge computing the Bebop drone is controlled by the PID controller following the predicted waypoints. 
The edge computer is an Alienware with the neural network fitted on the Nvidia GeForce GTX 960M and the rest of boiler code running on the Intel Core i7 CPU at 2.5GHz. The setup reaches 7 to 10fps. 
As the room was a bit dark, the brightness of the incoming frames are adjusted for all models to a constant value of 140 with PIL's Image Enhance function \cite{imageenhance}.

The desired trajectory has a spiral form with an increasingly sharp turn as shown in Fig. \ref{fig:wpreal}.
The white board is used to make sure the starting position of the drone is similar by taking of from the exact same position.
The baseline model and the deep supervision model clearly fail to follow at the end of the sharper turn, with the baseline model resulting in more variation at the end.
The deep supervision with triplet loss succeeds as only model five times the full trajectory.

As we did for the simulated setup, direct velocity prediction was compared with waypoints, shown in the right of Fig. \ref{fig:wpreal}. Again the DS + TL model with without abstraction was evaluated five times but clearly without success. This failure is on the one hand due to a poor dynamics model in simulation, but on the other hand also due to the much lower update rate. The PID controller uses an asynchronous Kalman filter compensating for disturbances at 30FPS instead of the slow 8FPS reached by the neural network on the GPU.

In order to push the limits of our model, we also train a neural network for following a red water tube and evaluate it in the real world above a grid of red lines, see Fig. \ref{fig:wpreal}. The only real difference between the foreground and background is the texture of the red water tube. The model could succeed about half of the times with maximally 3 successful turns in a row before getting distracted by the background.

% Eventually, the DS + TL model for blue line following was evaluated on various backgrounds where it succeeded at each test, flying over a white background, a toy carpet and grass.

%\thanks{$^1$Movie is available at https://drive.google.com/file/d/1OY_fxJP0Y35btvDy9My4DmitU4DsVysv/view?usp=sharing.}

%%%%%%%%%%%%%%%%%%%%%%%%%%%%%%%%%%%%%%%%%%%%%%%%%%%%%%%%%%%%%%%%%%%%%%%%%%%%%%%%
\section{Discussion}
\label{discussion}
% 0.25p
The experiments compared the impact of background augmentation, foreground augmentation, deep supervision; triplet loss regularisation and waypoint abstraction. With the use of the auxiliary mask prediction, we can successfully evaluate the visual generalisation capabilities of the encoder. With the use of deep supervision and the contrastive triplet loss, the mask prediction on validation data and real-world data was significantly improved.
Later the online performance was evaluated quantitatively in simulation, demonstrating similar good performance between the baseline and the more complex models. And finally the best models were compared on the real bebop with 5/5 success for our proposed DS + TL model with abstraction.

So as to conclude, this work introduced the idea of visual navigation with a visual cue in the foreground. With a good simulated model of this cue and simple background augmentation, a good waypoint prediction model can be trained in simulation and exploited in the real world without any training on real-world data. The use of a mask decoder gives insight in what relevant visual information is available in the feature used for decoding the waypoints. The quality of these masks can be boosted with deep supervision and a triplet loss.
Also the abstraction of velocities to waypoints was crucial to ensure a smooth transition in dynamics from simulation to the real world. The latter is probably caused by the much higher update rate of 30fps the PID controller with asynchronous kalman filter can achieve than the slow 10fps vision-based neural network.

These conclusions are only applicable for this scenario of following a line with a downward camera. Further research must investigate whether similar results are to be expected on different scenarios, such as flying through a gate, tracking a moving target, navigating in a lane. 

Rather than a full answer to sim2real, we propose this work as a new angle to look at the sim2real problem with new opportunities. 
The proposed techniques make the deep neural network focus more on what is relevant without loosing itself to misleading correlations and without relying on big labelled datasets, two complications that trouble many machine learning robotic researchers.


\bibliography{IEEEabrv, bibliography}

\end{document}